# Identifying Dynamic Sequential Plans

Jin Tian
Department of Computer Science
Iowa State University
Ames, IA 50011
*jtian@cs.iastate.edu*

## Abstract

We address the problem of identifying dynamic sequential plans in the framework of causal Bayesian networks, and show that the problem is reduced to identifying causal effects, for which there are complete identification algorithms available in the literature.

## 1 Introduction

This paper deals with the problem of evaluating the effects of sequential plans from a combination of nonexperimental data and qualitative causal assumptions. The causal assumptions will be represented in the form of an acyclic causal diagram [Spirtes et al., 1993, Heckerman and Shachter, 1995, Lauritzen, 2000, Pearl, 2000] in which arrows represent the potential existence of direct causal relationships between the corresponding variables. The causal diagram may contain unmeasured variables, and our task is to decide whether we can estimate the effects of a sequence of actions from the observed data.

We motivate the study by considering a medical treatment problem discussed in [Pearl and Robins, 1995, Dawid and Didelez, 2005]. There are a sequence of medical treatments $(X_1, \ldots, X_k)$ applied to a patient over time. We have observations $Z_i$ before and between the treatments. Doctors may prescribe a treatment based on previous treatments and observations. There is a outcome variable $Y$ (say survival) of special interest and we want to estimate the effects of the sequential treatments on $Y$. In general there may be unobserved confounders that have influence on the observed variables.

There are different possible strategies for choosing the treatment action $(X_i)$. The simplest action involves fixing the value of $X_i$ to a particular value $x_i^*$, called atomic intervention and denoted by $do(x_i^*)$ in [Pearl, 2000], e.g. fixing the dosage of a treatment irrespective of any observations on the patient. A unconditional plan consists of a sequence of pre-defined atomic actions. The problem of identifying unconditional plans is to compute the distribution of $Y$ under atomic interventions on a set $X$ of action variables, denoted by $P_x(y) = P(y|do(x))$, a quantity known as the causal effects of $X$ on $Y$. Sufficient graphical criteria for the identifiability of unconditional plans are derived in [Pearl and Robins, 1995]. Recently the general problem of identifying causal effects $P_x(y)$ in a causal diagram containing unobserved variables has been solved and complete algorithms for identification are given in [Tian and Pearl, 2003, Shpitser and Pearl, 2006b, Huang and Valtorta, 2006].

In general we may want to use dynamic treatment strategies in which the values of action variables $(X_i)$ are determined based on the previously observed variables and previously taken actions. Sufficient graphical criteria for the identifiability of dynamic plans are derived in [Dawid and Didelez, 2005]. However the identifiability problem is far from being solved.

In this paper, we show that the problem of identifying dynamic sequential plans can be reduced to the well-studied problem of identifying causal effects and therefore essentially solved the sequential plan identification problem. Although Pearl (2000, Section 4.2) has suggested that dynamic conditional plans may be identified by identifying conditional causal effects of the form $P_x(y|c)$, for which complete identification algorithms have been developed [Tian, 2004, Shpitser and Pearl, 2006a], in this paper, we will show that this gives a sufficient condition for identifying dynamic sequential plans but it is not necessary.

The rest of the paper is organized as follows. In Section 2, we review the work in [Dawid and Didelez, 2005] and define useful notation. In Section 3, we formulate the sequential plan problem in the framework of causal Bayesian networks. We show how to reduce the problem of identifying dynamic sequential plans into a problem of identifying causal effects in Section 4, and discuss in Section 5 the problem versus that of identifying unconditional plans and conditional causal effects. Section 6 concludes the paper.

## 2 Previous Work and Notation

Dawid and Didelez (2005) formulated the problem of identifying dynamic sequential plans in the framework of regime indicators and influence diagrams. An influence diagram (ID) is a DAG over a set $V = \{V_1, \ldots, V_n\}$ of variables that also includes regime indicators as special nodes of their own called decision nodes [Dawid, 2002]. We assume that all variables are discrete. The DAG is assumed to represent conditional independence assertions that each variable is independent of all its non-descendants given its direct parents in the graph.[1] These assertions imply that the joint probability function $P(v) = P(v_1, \ldots, v_n)$ factorizes according to the product [Pearl, 1988]

$$P(v) = \prod_i P(v_i|pa_i) \quad (1)$$

where $pa_i$ are (values of) the parents of variable $V_i$ in the graph.[2]

The question of causal inference is considered as a problem of inference across different *regimes*, in which we may intervene in certain variables in certain ways and observe the behavior of other variables. *Regime indicators* are used to represent different types of interventions. Here we will roughly follow the notation used in [Didelez *et al.*, 2006]. The regime indicator for an intervention in a variable $V_i$ is denoted by $\sigma_{V_i}$ and can take values in a set of strategies. Under strategy $\sigma_{V_i}$, the conditional probability $P(v_i|pa_i)$ is changed to $P(v_i|pa_i; \sigma_{V_i})$. We will consider the following types of interventions.

- *Idle regime* $\sigma_{V_i} = \emptyset$: No intervention takes place, therefore

  $$P(v_i|pa_i; \sigma_{V_i} = \emptyset) = P(v_i|pa_i).$$

  The idle regime is also called the observational regime under which we will assume that observational data has been collected. Therefore $P(v_i|pa_i)$ can be estimated from data if $V_i$ and $Pa_i$ are observed.

- *Atomic intervention* $\sigma_{V_i} = do(v_i^*)$: The strategy of setting $V_i$ to a fixed value $v_i^*$, denoted by $do(V_i = v_i^*)$ or simply $do(v_i^*)$ in Pearl (2000), such that

  $$P(v_i|pa_i; \sigma_{V_i} = do(v_i^*)) = \delta(v_i, v_i^*),$$

  where $\delta(v_i, v_i^*)$ is one if $v_i = v_i^*$ and zero otherwise.

- *Conditional intervention* $\sigma_{V_i} = do(g(c))$: In general, $V_i$ may be made to respond in a specified way to some set $C$ of previously observed variables, denoted by $do(V_i = g(c))$ in Pearl (2000), such that

  $$P(v_i|pa_i; \sigma_{V_i} = do(g(c))) = \delta(v_i, g(c)),$$

  where $g(.)$ is a pre-specified deterministic function and the variables in $C$ can not be descendants of $V_i$.

- *Random intervention* $\sigma_{V_i} = d_C$: More generally, we may let $V_i$ take on a random value according to some distribution possibly depending on some set $C$ of previously observed variables such that

  $$P(v_i|pa_i; \sigma_{V_i} = d_C) = P^*(v_i|c),$$

  where $P^*(v_i|c)$ is a pre-specified probability distribution and the variables in $C$ can not be descendants of $V_i$.

In a sequential decision problem, we may intervene, at least in principle, in a set of variables $X = \{X_i\} \subset V$, called *control variables* or *action variables*, and are interested in the response of a variable $Y$, called *response variable* or *outcome variable*. Let $Z$ be the rest of observed variables which are often called *covariates*. The variables are assumed to be ordered in a sequence $(L_1, X_1, \ldots, L_K, X_K, Y)$ where $L_i \subseteq Z$ are the set of observed covariates after $X_{i-1}$ and before $X_i$. We denote $\bar{L}_i = (L_1, \ldots, L_i)$ and $\bar{X}_i = (X_1, \ldots, X_i)$.

Given an intervention strategy $\sigma_X = \{\sigma_{X_i}\}$, under a condition called *simple stability* which says that the observed covariates $L_i$ and the outcome $Y$ are independent of how action variables are generated once all earlier observables ($\bar{L}_{i-1}, \bar{X}_{i-1}$ for $L_i$; $X, Z$ for $Y$) are given, Dawid and Didelez (2005) show that the post-intervention distribution of $Y$ is identified as

$$P(y; \sigma_X) = \sum_{x,z} P(y|x,z) \prod_i P(l_i|\bar{l}_{i-1}, \bar{x}_{i-1})$$
$$\prod_i P(x_i|\bar{x}_{i-1}, \bar{l}_i; \sigma_{X_i}), \quad (2)$$

where $P(x_i|\bar{x}_{i-1}, \bar{l}_i; \sigma_{X_i})$ are determined by the chosen regime and the other quantities can be estimated from observational data. Eq. (2) is known as the *G-formula*, and has been obtained in [Robins, 1986, Robins, 1987] in the framework of potential response models.

---

[1] We use family relationships such as "parents," "children," and "ancestors" to describe the obvious graphical relationships.

[2] We use uppercase letters to represent variables or sets of variables, and use corresponding lowercase letters to represent their values (instantiations).

When there are unobserved confounders, the simple stability may not hold. Dawid and Didelez (2005) makes *extended stability* assumption which essentially is (simple) stability with respect to the extended domain that includes unobserved $U$ variables ignoring the distinction between $Z$ and $U$. The G-formula (2) no longer holds unless we include unobserved $U$ variables, but then the conditional probabilities involving $U$ variables can no longer be estimated from the data. Sufficient graphical criteria for identifying $P(y; \sigma_X)$ are derived. The criteria were obtained by identifying graphical conditions under which the simple stability can be regained such that the G-formula can be used, and by extending the work in [Pearl and Robins, 1995] to dynamic plans.

## 3 Problem Formulation

In this paper, we will formulate the sequential plan problem in the framework of causal Bayesian networks. A *causal Bayesian network (CBN)* consists of a DAG $G$ over a set $V = \{V_1, \ldots, V_n\}$ of variables, called a *causal diagram*. The interpretation of such a graph has two components, probabilistic and causal. The probabilistic interpretation views $G$ as representing conditional independence assertions such that the joint probability function $P(v) = P(v_1, \ldots, v_n)$ factorizes according to Eq. (1). The causal interpretation views the directed edges in $G$ as representing causal influences between the corresponding variables. In this interpretation, the factorization of (1) still holds, but the factors are further assumed to represent autonomous data-generation processes, that is, each conditional probability $P(v_i|pa_i)$ represents a stochastic process by which the values of $V_i$ are chosen in response to the values $pa_i$ (previously chosen for $V_i$'s parents), and the stochastic variation of this assignment is assumed independent of the variations in all other assignments. Moreover, each assignment process remains invariant to possible changes in the assignment processes that govern other variables in the system. This modularity assumption enables us to predict the effects of interventions, whenever interventions are described as specific modifications of some factors in the product of (1). We typically assume that every variable $V_i$ can potentially be manipulated by external intervention. So we might think of a CBN as an ID such that each node is (implicitly) pointed to by a corresponding regime/intervention indicator.

In a sequential decision problem, we may intervene in a set of action variables $X = \{X_i\} \subset V$, and are interested in the response of a set of outcome variables $Y$. Assume that all the variables $V$ are observed and let the rest of covariate variables be $Z = \{Z_i\} = V \setminus (X \cup Y)$. Given an intervention strategy $\sigma_X = \{\sigma_{X_i}\}$, by modularity assumption, we can predict the effects of $\sigma_X$ as

$$P(v; \sigma_X)$$
$$= \prod_i P(y_i|pa_{y_i}) \prod_i P(z_i|pa_{z_i}) \prod_i P(x_i|pa_{x_i}; \sigma_{X_i}), \quad (3)$$

where, by modularity assumption, those conditional probabilities corresponding to unmanipulated variables remain unaltered. We note that Dawid and Didelez's (2005) simple stability assumption leads to Eq. (3) in the framework of CBNs. We see that, given a CBN, whenever all variables in $V$ are observed, the consequence of an intervention strategy on the outcome variables $Y$ is computed as

$$P(y; \sigma_X) = \sum_{x,z} \prod_i P(y_i|pa_{y_i}) \prod_i P(z_i|pa_{z_i})$$
$$\prod_i P(x_i|pa_{x_i}; \sigma_{X_i}), \quad (4)$$

where $P(x_i|pa_{x_i}; \sigma_{X_i})$ are determined by the chosen regime and the other quantities can be estimated from observational data. We note that the G-formula (2) can be reduced to Eq. (4) by using the conditional independence relationships implied by the CBN that each variable is independent of all its non-descendants given its parents.

In general we may be concerned with confounding effects due to unobserved influential variables. In the presence of unobserved confounders, the distribution over observed variables can no longer factorize according to (1). Letting $V = Y \cup Z \cup X$ and $U = \{U_1, \ldots, U_{n'}\}$ stand for the sets of observed and unobserved variables, respectively, the observed probability distribution, $P(v)$, becomes a mixture of products:

$$P(v) = \sum_u \prod_{\{i|V_i \in V\}} P(v_i|pa_{v_i}) \prod_{\{i|U_i \in U\}} P(u_i|pa_{u_i}). \quad (5)$$

We still make modularity assumption in the CBN with unobserved variables, and the effects of an intervention strategy $\sigma_X$ on the outcome variables $Y$ can be expressed as

$$P(y; \sigma_X) = \sum_{x,z,u} \prod_i P(y_i|pa_{y_i}) \prod_i P(z_i|pa_{z_i})$$
$$\prod_i P(x_i|pa_{x_i}; \sigma_{X_i}) \prod_i P(u_i|pa_{u_i}). \quad (6)$$

We note that Dawid and Didelez's (2005) extended stability assumption leads to Eq. (6) in the framework of CBNs. In (6), the quantities $P(y_i|pa_{y_i})$ and $P(z_i|pa_{z_i})$ (and $P(u_i|pa_{u_i})$) may involve elements of $U$ and may not be estimable from data. Then the question of identifiability arises, i.e., whether it is possible to express $P(y;\sigma_X)$ as a function of the observed distribution $P(v)$.

**Definition 1** [*Plan Identifiability*]
*A sequential plan is said to be identifiable if $P(y;\sigma_X)$ is uniquely computable from the observed distribution $P(v)$.*

## 4 Identification of Sequential Plans

First we make the following assumption about the type of interventions we will consider.

**Assumption 1** *$P(x_i|pa_{x_i};\sigma_{X_i})$ does not depend on the unobserved variable. That is, for conditional intervention $\sigma_{X_i} = do(g(c))$ or random intervention $\sigma_{X_i} = d_C$, we require $C \subseteq X \cup Z$.*

This assumption corresponds to Condition 6.6 or 7.2 in [Dawid and Didelez, 2005].

Under Assumption 1, Eq. (6) becomes

$$P(y;\sigma_X) = \sum_{x,z} \prod_i P(x_i|pa_{x_i};\sigma_{X_i}) \sum_u \prod_i P(y_i|pa_{y_i})$$
$$\prod_i P(z_i|pa_{z_i}) \prod_i P(u_i|pa_{u_i}) \qquad (7)$$
$$= \sum_{x,z} \prod_i P(x_i|pa_{x_i};\sigma_{X_i}) P_x(y,z) \qquad (8)$$

Obviously a sufficient condition for $P(y;\sigma_X)$ being identifiable is that the causal effect $P_x(y,z)$ is identifiable. In particular, a simple sufficient condition for $P_x(y,z)$ being identifiable is if all the parents of action $(X)$ variables are observables, which is Condition 6.3 in [Dawid and Didelez, 2005].

**Proposition 1** *If all the parents of action $(X)$ variables are observables, then $P(y;\sigma_X)$ is identifiable [Dawid and Didelez, 2005].*

*Proof:* If all the parents of action $(X)$ variables are observables, then $P(x_i|pa_{x_i})$ contains no unobserved $(U)$ variables, and Eq. (5) can be written as

$$P(v) = \prod_i P(x_i|pa_{x_i}) P_x(y,z), \qquad (9)$$

from which we obtain that $P_x(y,z)$ is identified as

$$P_x(y,z) = \frac{p(x,y,z)}{\prod_i P(x_i|pa_{x_i})} \qquad (10)$$
$$= \prod_{\{i|V_i \in Y \cup Z\}} P(v_i|\bar{v}_i), \qquad (11)$$

where we have used the chain rule assuming an order of $V$ variables that is consistent with the DAG and $\bar{v}_i$ denotes the $V$ variables ordered ahead of $V_i$. Hence the sequential plan is identified as

$$P(y;\sigma_X) = \sum_{x,z} \prod_i P(x_i|pa_{x_i};\sigma_{X_i}) \prod_{\{i|V_i \in Y \cup Z\}} P(v_i|\bar{v}_i), \qquad (12)$$

which is essentially the G-formula (2). □

In general $P_x(y,z)$ being identifiable is not a necessary condition for $P(y;\sigma_X)$ being identifiable. Eq. (7) may be simplified in that a factor $P(z_i|pa_{z_i})$ may be summed out (using $\sum_{z_i} P(z_i|pa_{z_i}) = 1$) if $Z_i$ does not appear in any other factors (graphically, if $Z_i$ does not have any children). We can derive stronger identification criterion by summing out as many factors as possible from Eq. (7). Before presenting our result, we first introduce some notation.

Following [Tian and Pearl, 2003], for any observed set $S \subseteq V$ of variables, we define the quantity $Q[S]$ to denote the post-intervention distribution of $S$ under atomic interventions to all other variables:

$$Q[S](v) = P_{v \setminus s}(s)$$
$$= \sum_u \prod_{\{i|V_i \in S\}} P(v_i|pa_{v_i}) \prod_{\{i|U_i \in U\}} P(u_i|pa_{u_i}). \qquad (13)$$

For convenience, we will often write $Q[S](v)$ as $Q[S]$. Eq. (7) can be written as

$$P(y;\sigma_X) = \sum_{x,z} \prod_i P(x_i|pa_{x_i};\sigma_{X_i}) Q[Y \cup Z]. \qquad (14)$$

Let $G_{\sigma_X}$ denote *the manipulated graph* under the intervention strategy $\sigma_X$, which can be constructed from the original causal graph $G$ as follows:

- For an atomic intervention $\sigma_{X_i} = do(x_i)$, cut off all the arrows entering $X_i$;

- For a conditional intervention $\sigma_{X_i} = do(g_i(c_i))$ or a random intervention $\sigma_{X_i} = d_{C_i}$, cut off all the arrows entering $X_i$ and then add an arrow entering $X_i$ from each variable in $C_i$.

Based on Eq. (14), we obtain the following sufficient criterion for identifying $P(y; \sigma_X)$.

**Theorem 1** *Let $Z_D$ be the set of variables in $Z$ that are ancestors of $Y$ in $G_{\sigma_X}$. $P(y; \sigma_X)$ is identifiable if the causal effects $Q[Y \cup Z_D] = P_{x, z \setminus z_D}(y, z_D)$ is identifiable.*

*Proof:* Let $X_D$ be the set of variables in $X$ that are ancestors of $Y$ in $G_{\sigma_X}$. Then all the non-ancestors of $Y$ can be summed out from Eq. (14) leading to

$$P(y; \sigma_X) = \sum_{x_D, z_D} \prod_{\{i|X_i \in X_D\}} P(x_i|pa_{x_i}; \sigma_{X_i}) Q[Y \cup Z_D] \quad (15)$$

$$= \sum_{x_D, z_D} \prod_{\{i|X_i \in X_D\}} P(x_i|pa_{x_i}; \sigma_{X_i}) P_{x, z \setminus z_D}(y, z_D). \quad (16)$$

□

We conjecture that the condition in Theorem 1 is also necessary. It might appear that Eq. (15) can be further simplified as follows. Let $Z_{\sigma_{X_D}}$ be the set of $Z$ variables that appear in the term $\prod_{\{i|X_i \in X_D\}} P(x_i|pa_{x_i}; \sigma_{X_i})$ (the set of conditioning variables in the strategy $\sigma_{X_D}$). Then the term $\prod_{\{i|X_i \in X_D\}} P(x_i|pa_{x_i}; \sigma_{X_i})$ is a function of $X_D$ and $Z_{\sigma_{X_D}}$. Eq. (15) becomes

$$P(y; \sigma_X)$$
$$= \sum_{x_D, z_{\sigma_{X_D}}} \prod_{\{i|X_i \in X_D\}} P(x_i|pa_{x_i}; \sigma_{X_i}) \sum_{z_D \setminus z_{\sigma_{X_D}}} Q[Y \cup Z_D] \quad (17)$$

$$\equiv \sum_{x_D, z_{\sigma_{X_D}}} g(x_D, z_{\sigma_{X_D}}) \sum_{z_D \setminus z_{\sigma_{X_D}}} Q[Y \cup Z_D] \quad (18)$$

From Eq. (18), $P(y; \sigma_X)$ is identifiable if $\sum_{z_D \setminus z_{\sigma_{X_D}}} Q[Y \cup Z_D]$ is identifiable, and intuitively, the condition appears to be necessary too, since the term $g(x_D, z_{\sigma_{X_D}})$ is specified externally and no more factors can be summed out (as far as the function $g(.)$ is not independent of any variables in $z_{\sigma_{X_D}}$). A strict proof of this necessity is still under study.

On the other hand, due to the fact that none of the factors corresponding to the variables in $Z_D \setminus Z_{\sigma_{X_D}}$ can be summed out from $Q[Y \cup Z_D]$, it has been shown that $\sum_{z_D \setminus z_{\sigma_{X_D}}} Q[Y \cup Z_D]$ can be identified only via identifying $Q[Y \cup Z_D]$ and it is a if and only if condition (Lemma 11 in [Huang and Valtorta, 2006]). So from the point of view of identifying $P(y; \sigma_X)$ the reduction from Eq. (15) to Eq. (17) is not necessary.

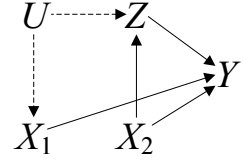

Figure 1: An example causal graph

We therefore have reduced the problem of identifying dynamic sequential plans $P(y; \sigma_X)$ into that of identifying causal effects $Q[Y \cup Z_D]$ while the latter problem has been solved and complete algorithms are given in [Tian and Pearl, 2003, Shpitser and Pearl, 2006b, Huang and Valtorta, 2006].

We demonstrate the application of Theorem 1 and the identification process with an example. Consider the problem of identifying $P(y; \sigma_{X_1}, \sigma_{X_2})$ in Figure 1, which was studied in [Dawid and Didelez, 2005] and is troubling to the methods presented therein. Theorem 1 calls for identifying $Q[\{Y, Z\}]$ which can be shown to be identifiable. We are given the observational distribution

$$P(v) = P(y|x_1, x_2, z) P(x_2) Q[\{Z, X_1\}], \quad (19)$$

where

$$Q[\{Z, X_1\}] = \sum_u P(z|x_2, u) P(x_1|u) P(u). \quad (20)$$

We want to compute

$$P(y; \sigma_{X_1}, \sigma_{X_2})$$
$$= \sum_{x_1, x_2, z} P(x_1; \sigma_{X_1}) P(x_2; \sigma_{X_2}) P(y|x_1, x_2, z) Q[\{Z\}], \quad (21)$$

which calls for computing $Q[\{Z\}]$. From Eq. (20), it is clear that

$$Q[\{Z\}] = \sum_{x_1} Q[\{Z, X_1\}]. \quad (22)$$

From Eq. (19), we obtain

$$Q[\{Z, X_1\}] = P(z, x_1|x_2) \quad (23)$$

Therefore $Q[\{Z\}]$ is identified and we finally obtain

$$P(y; \sigma_{X_1}, \sigma_{X_2})$$
$$= \sum_{x_1, x_2, z} P(x_1; \sigma_{X_1}) P(x_2; \sigma_{X_2}) P(y|x_1, x_2, z) P(z|x_2). \quad (24)$$

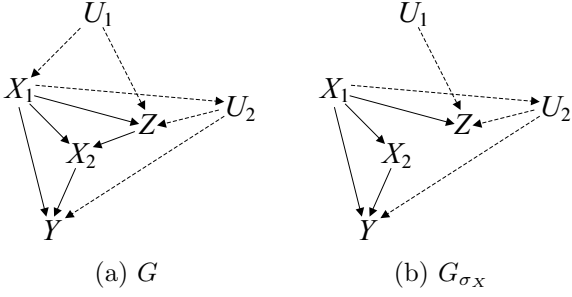

(a) $G$  (b) $G_{\sigma_X}$

Figure 2: An example causal graph

## 5 Discussion

### 5.1 Unconditional plans are easier

In general identifying dynamic plans is more difficult than identifying unconditional plans that involve only atomic interventions. Let an intervention strategy $\sigma'_X$ consist of all atomic interventions. Then the manipulated graph $G_{\sigma'_X}$ is a subgraph of $G_{\sigma_X}$. Let $Z'_D$ be the set of variables in $Z$ that are ancestors of $Y$ in $G_{\sigma'_X}$. Then $Z'_D$ is a subset of $Z_D$. We have

$$P(y; \sigma'_X) = P_x(y) = \sum_{z'_D} Q[Y \cup Z'_D]. \quad (25)$$

Therefore identifying $P_x(y)$ calls for identifying $Q[Y \cup Z'_D]$ while $P(y; \sigma_X)$ calls for identifying $Q[Y \cup Z_D]$. Now we have

$$Q[Y \cup Z'_D] = \sum_{z_D \setminus z'_D} Q[Y \cup Z_D] \quad (26)$$

The factors of the variables in $Z_D \setminus Z'_D$ are summed out from $Q[Y \cup Z_D]$ since the variables in $Z_D \setminus Z'_D$ can be ancestors of $Y$ only through $X_i$'s. In general, whenever $Q[Y \cup Z_D]$ (and therefore $P(y; \sigma_X)$) is identifiable, then $P_x(y)$ is identifiable. However, it is possible that $P_x(y)$ is identifiable but $Q[Y \cup Z_D]$ (and therefore $P(y; \sigma_X)$) is not.

We demonstrate this point with an example. Consider the problem of identifying $P(y; \sigma_{X_1}, \sigma_{X_2})$ in Figure 2(a), which was studied in [Pearl and Robins, 1995]. If $P(x_2|x_1, z; \sigma_{X_2})$ depends on $Z$, say $\sigma_{X_2} = do(g(x_1, z))$, then $Z_D = \{Z\}$, and by Theorem 1, to identify $P(y; \sigma_{X_1}, \sigma_{X_2})$ we need to identify $Q[\{Y, Z\}]$, which can be shown to be not identifiable (by theorems in [Huang and Valtorta, 2006]). More specifically, given the observational distribution

$$P(v) = P(x_2|x_1, z)Q[\{X_1, Z, Y\}], \quad (27)$$

we want to identify

$$P(y; \sigma_{X_1}, \sigma_{X_2}) = \sum_{x_1, x_2, z} P(x_1; \sigma_{X_1})P(x_2|x_1, z; \sigma_{X_2})Q[\{Y, Z\}] \quad (28)$$

From Eq. (28), we see that if $P(x_2|x_1, z; \sigma_{X_2})$ depends on $Z$, then the identifiability of $P(y; \sigma_{X_1}, \sigma_{X_2})$ depends on the identifiability of $Q[\{Y, Z\}]$. We therefore conclude that $P(y; \sigma_{X_1}, \sigma_{X_2})$ is not identifiable.

On the other hand, if $P(x_2|x_1, z; \sigma_{X_2})$ is independent of $Z$, say $P(x_2|x_1, z; \sigma_{X_2}) = P^*(x_2|x_1)$ (or $\sigma_{X_2} = do(x_2)$), then the set $Z_D$ of variables in $Z$ that are ancestors of $Y$ in $G_{\sigma_X}$ becomes empty (see Figure 2(b)), and, by Theorem 1, the identifiability of $P(y; \sigma_{X_1}, \sigma_{X_2})$ depends on the identifiability of $Q[\{Y\}]$. In fact, in this case, Eq. (28) becomes

$$P(y; \sigma_{X_1}, \sigma_{X_2} = d_{X_1})$$
$$= \sum_{x_1, x_2} P(x_1; \sigma_{X_1}) P^*(x_2|x_1) \sum_z Q[\{Y, Z\}]$$
$$= \sum_{x_1, x_2} P(x_1; \sigma_{X_1}) P^*(x_2|x_1) Q[\{Y\}] \quad (29)$$

From Eq. (27) we obtain

$$Q[\{X_1, Z, Y\}] = P(v)/P(x_2|x_1, z)$$
$$= P(y|x_1, x_2, z)P(x_1, z). \quad (30)$$

It can be shown (or confirmed) that

$$Q[\{Y\}] = \frac{\sum_z Q[\{X_1, Z, Y\}]}{\sum_{y, z} Q[\{X_1, Z, Y\}]}$$
$$= \sum_z P(y|x_1, x_2, z)P(z|x_1). \quad (31)$$

We obtain

$$P(y; \sigma_{X_1}, \sigma_{X_2} = d_{X_1})$$
$$= \sum_{x_1, x_2} P(x_1; \sigma_{X_1}) P^*(x_2|x_1) \sum_z P(y|x_1, x_2, z)P(z|x_1). \quad (32)$$

And in particular, the unconditional plan is identified as

$$P_{x_1, x_2}(y) = Q[\{Y\}] = \sum_z P(y|x_1, x_2, z)P(z|x_1). \quad (33)$$

### 5.2 Identification via conditional causal effects?

Pearl (2000) has suggested that dynamic sequential plans involving conditional and random interventions may be identified by identifying conditional causal effects of the form $P_x(y|z)$. For interventions on a single variable $X$, we can show [Pearl, 2000, Section 4.2] that

$$P(y; \sigma_X = do(g(z))) = \sum_z P_x(y|z)|_{x=g(z)} P(z),$$

and

$$P(y; \sigma_X = d_Z) = \sum_{x,z} P_x(y|z) P^*(x|z) P(z).$$

Therefore it appears that $P(y; \sigma_X)$ is identifiable if and only if $P_x(y|z)$ is identifiable.

This idea was generalized to dynamic sequential plans. Consider a plan involving a sequence of conditional interventions $\sigma_{X_i} = do(g_i(C_i))$. Let $Z_{\sigma_X} = Z \cap (\cup_i C_i)$ be the set of conditioning variables in the strategy $\sigma_X$. Pearl (2006) shows that

$$P(y; \sigma_X) = \sum_{z_{\sigma_X}} P_{x_z}(y|z_{\sigma_X}) P_{x_z}(z_{\sigma_X}), \qquad (34)$$

where $x_z$ are the values attained by $X$ when the conditioning set $Z_{\sigma_X}$ takes the values $z_{\sigma_X}$. Pearl then suggests that sequential conditional plans can be identified by identifying conditional causal effects $P_x(y|z)$ and $P_x(z)$. This motivated the study of the identifiability of conditional causal effects and complete algorithms have been developed in [Tian, 2004, Shpitser and Pearl, 2006a].

Next we show that although the identification of $P_{x_z}(y|z_{\sigma_X})$ and $P_{x_z}(z_{\sigma_X})$ is sufficient for identifying $P(y; \sigma_X)$, it is nonetheless not necessary. Rewrite Eq. (34) in the following

$$P(y; \sigma_X) = \sum_{x, z_{\sigma_X}} \delta(x_i, g_i(C_i)) P_x(y, z_{\sigma_X}) \qquad (35)$$

$$= \sum_{x, z_{\sigma_X}} \delta(x_i, g_i(C_i)) \sum_{z \setminus z_{\sigma_X}} Q[Y, Z] \qquad (36)$$

Comparing the reduction from Eq. (14) into (17) with the reduction from (14) to (36), we obtain that if $X_D = X$ then (36) is equivalent to (17), otherwise Eq. (36) may be further reduced in that more factors could be summed out from $Q[Y, Z]$. We obtain the following conclusion

- If all the variables in $X$ are ancestors of $Y$ in $G_{\sigma_X}$, then the sequential plan $P(y; \sigma_X)$ can be identified by identifying the causal effects $P_x(y, z_{\sigma_X})$, otherwise it is possible that $P(y; \sigma_X)$ is identifiable even if $P_x(y, z_{\sigma_X})$ is not.

We demonstrate this point with an example. Consider the problem of identifying $P(y; \sigma_X)$ where $\sigma_X = \{\sigma_{X_1} = do(g_1(Z_1)), \sigma_{X_2} = do(g_2(Z_2)), \sigma_{X_3} = do(g_3(Z_3))\}$ in Figure 3(a). The graph $G_{\sigma_X}$ is shown in Figure 3(b). We have $Z_D = \{Z_1, Z_3\}$, and Theorem 1 calls for identifying $Q[\{Y, Z_1, Z_3\}]$ which can be shown to be identifiable. On the other hand, $P_{x_1 x_2 x_3}(y, z_1, z_2, z_3) = Q[\{Y, Z_1, Z_2, Z_3\}]$ is not identifiable. More specifically, given the observational distribution

$$P(v) = P(y|x_1, x_3, z_3) P(x_3|x_2, z_3) P(z_1) Q[\{X_2, Z_3\}] Q[\{X_1, Z_2\}], \qquad (37)$$

we want to identify

$$P(y; \sigma_X)$$
$$= \sum_{x_i, z_i} \prod_i \delta(x_i, g_i(z_i)) P(y|x_1, x_3, z_3) P(z_1)$$
$$\qquad Q[\{Z_3\}] Q[\{Z_2\}] \qquad (38)$$
$$= \sum_{x_1, x_3, z_1, z_3} \delta(x_1, g_1(z_1)) \delta(x_3, g_3(z_3)) P(y|x_1, x_3, z_3)$$
$$\qquad P(z_1) Q[\{Z_3\}], \qquad (39)$$

where $Q[\{Z_3\}]$ can be identified as

$$Q[\{Z_3\}] = \sum_{u_2} P(z_3|u_2) P(u_2) = P(z_3). \qquad (40)$$

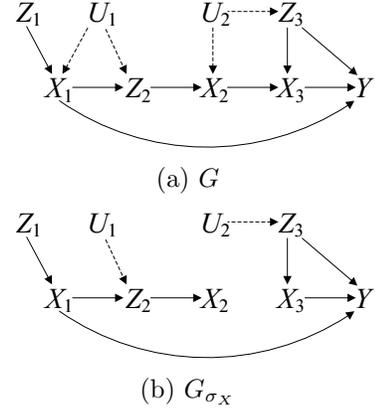

Figure 3: An example causal graph

We obtain

$$P(y; \sigma_X) = \sum_{z_1, z_3} P(y|g_1(z_1), g_3(z_3), z_3) P(z_1) P(z_3). \qquad (41)$$

On the other hand,

$$P_{x_1 x_2 x_3}(y, z_1, z_2, z_3)$$
$$= P(y|x_1, x_3, z_3) P(z_1) Q[\{Z_3\}] Q[\{Z_2\}]$$
$$= P(y|x_1, x_3, z_3) P(z_1) P(z_3) Q[\{Z_2\}] \qquad (42)$$

is not identifiable since $Q[\{Z_2\}]$ is not identifiable. In fact the conditional causal effect

$$P_{x_1x_2x_3}(y|z_1, z_2, z_3) = P(y|x_1, x_3, z_3) \quad (43)$$

is identifiable but the causal effect

$$P_{x_1x_2x_3}(z_1, z_2, z_3) = P(z_1)P(z_3)Q[\{Z_2\}] \quad (44)$$

is not identifiable.

## 6 Conclusion

We present a method for identifying dynamic sequential plans. A closed-form expression for the probability of the outcome variables under a dynamic plan can be obtained in terms of the observed distribution, by using the algorithms for identifying causal effects available in the literature.

### Acknowledgments

This research was partly supported by NSF grant IIS-0347846.